# *Shamap*: Shape-based Manifold Learning

Fenglei Fan, Ziyu Su, Yueyang Teng, Ge Wang*

Biomedical Imaging Center, BME/CBIS, ECSE

Rensselaer Polytechnic Institute, Troy, New York, USA

fanf2@rpi.edu, suz4@rpi.edu, tengyy.bmie@gmail.com, ge-wang@ieee.org

*Abstract*—**For manifold learning, it is assumed that high-dimensional sample/data points are embedded on a low-dimensional manifold. Usually, distances among samples are computed to capture an underlying data structure. Here we propose a metric according to angular changes along a geodesic line, thereby reflecting the underlying shape-oriented information or a topological similarity between high- and low-dimensional representations of a data cloud. Our results demonstrate the feasibility and merits of the proposed dimensionality reduction scheme.**

## I. INTRODUCTION

High dimensional big data，like speech signals, image data，or medical records, are frequently tackled in many practical applications. Especially, for machine learning research the dimensionality of data will directly affect the computational cost in training and testing a deep neural network. Moreover, high dimensional data usually contain a significant amount of noise and artifacts, which may mislead the model in solving classification and regression tasks [1]. Importantly, in many cases high-dimensional data are essentially a low-dimensional embedding in an ambient space. In other words, a data cloud lives on a low-dimensional manifold embedded in the high dimensional space [2]. This key observation suggests the possibility of dimensionality reduction to facilitate visualization and analysis of the intrinsic structure of involved data.

Initially, linear dimensionality reduction techniques, like principal component analysis (PCA) [3] and multi-dimensional scaling (MDS) [4], are mainstream methods for dimensionality reduction. It is interesting that PCA focuses on the maximum variance between data when projecting data from a high dimensional space to a low dimensional space, while MDS keeps the original distances between data. However, in most challenging problems, the intrinsic structures of datasets are usually nonlinear. Linear methods would suffer from serious overlapping or aliasing issues, since the nonlinear distribution of data in the original space is not fully taken into account [5].

After the initial progress made in 2000, manifold learning became one of the mainstream nonlinear dimensionality reduction techniques [6]. Driven by major academic curiosities and real-world needs, novel algorithms, such as Isomap [2], Locally Linear Embedding (LLE) [7]，Laplacian eigenmap [8] and kernel-Principal Component Analysis (k-PCA) [9], were developed to flatten a convoluted manifold by its intrinsic structure.  As one of the representative methods, Isomap combines the Floyd-Warshall algorithm [10] with MDS to compress high-dimensional data. It feeds geodesic distances into the MDS framework to maintain relationship among data in their neighborhood. In contrast to Isomap, LLE imposes the constraint that data are locally linear and manages to find a similar linear combination between data in low dimensions as in the original space. Laplacian eigenmap is a variant of LLE, which maps data to a low dimension representation by applying eigen decomposition to the graph Laplacian matrix [11] without changing the intrinsic configuration of data. k-PCA was proposed to compensate for the drawback of regular PCA, which uses the kernel trick to nonlinearly map the data into a kernel space. Although these methods are successful in keeping the metric relations, none of these or any other existing algorithm pays attention on the shape of a manifold which contains a data distribution.

Inspired by Isomap, we are interested in shape-based manifold learning; i.e., our goal is to map a high-dimensional data cloud to a low-dimensional counterpart wherein the shapes of the two presentations are similar. In contrast to Isomap that incorporates the geodesic distance into MDS, what we propose here is called Shamap as an angularly oriented version of MDS. In other words, while Isomap is distance-wise specific, Shamap is angularly sensitive. Indeed, angular relations are a significant aspect of data structures. For example, a polar coordinate system is often more meaningful and convenient in representing many important curves than a Cartesian coordinate system. Since the angular increment is perpendicular to the tangential direction of a trajectory, the angular representation is most suitable to capture the shape of the trajectory.

Specifically, we propose to replace the geodesic distance with the accumulated angular changes along a geodesic line in the framework of Isomap to form our proposed Shamap algorithm. We compute the angle between two neighboring vectors according to the following formula:

$$\cos(\theta_{ij}) = \frac{(x_i - c)^T(x_j - c)}{||x_i - c|| \cdot ||x_j - c||}. \tag{1}$$

Unlike k-PCA, which employs a kernel function to map the data cloud into a kernel space and then apply PCA techniques, Isomap and Shamap are based on MDS. As we discussed above, Isomap and Shamap are good at maintaining manifold's original geodesic distance and angular difference in a low dimensional space. This attribute can be utilized to facilitate supervised and unsupervised learning tasks by carrying out the learning process in a low dimensional space without any significant information loss nor performance tradeoff.

While local embedding methods can also keep the global structure of a data cloud, such as Locally Linear Embedding (LLE) and Hessian LLE [12], both Isomap and Shamap compute global features directly in terms of geodesic distance and geodesic tangential change respectively. At the same time, iterative optimization is not needed in Isomap and Shamap, unlike those local methods. Due to the shape preserving property of Shamap, it is potentially more advantageous in unraveling convoluted structures than Isomap.

## II. SHAMAP

Given a dataset comprising of $N$ instances $x_1, x_2, x_3 \ldots, x_N$, the tangential angles at those points are calculated with Eq. (1). Specifically, the proposed Shamap algorithm consists of the following three steps:

**#1 Find K-nearest neighbors:** Determine whether a pair of points $i$ and $j$ are connected or not according to the K-nearest neighbor (KNN) criterion, or if they are within a fixed distance $\epsilon$.

**#2 Compute the total angular change:** To compute the total angular change $\theta_{mn}$, between two points $m$ and $n$, we first determine the shortest path connecting data indexed by $m$ and $n$, such as in the sequence of indices $(m, k_1, k_2, \ldots, k_W, n)$, then $\theta_{mn}$ is derived by accumulating incremental angular changes along the geodesic line:

$$\theta_{mn} = \theta_{mk_1} + \theta_{k_1 k_2} + \theta_{k_2 k_3} + \cdots + \theta_{k_W n} \tag{2}$$

where pairs of successive subscripts $(m, k_1), (k_1, k_2) \ldots (k_W, n)$ denote pairs of neighboring points.

**#3 Construct a $d$-dimensional embedding:** Compute cosine values of local angular changes to form a matrix $C$, which can round those cumulative angular differences that exceed one circle period periodically. Let $\lambda_p$ be the $p^{th}$ eigenvalue of the matrix $C$, and $u_i^p$ be the $i^{th}$ component of the $p^{th}$ eigenvector. Then, the $p^{th}$ component of the $d$-dimensional coordinate vector $y^p$ is computed as:

$$y^p = \sqrt{\lambda_p} u_i^p \|x_i - c\| \qquad (3)$$

The scaling factor $\|x_i - c\|$ is needed to reasonably project the high-dimensional points onto a low-dimensional manifold according to the vector norm relative to the reference point $c$.

To recap the implementation details of Shamap, the pseudo-code is given as follows:

---
**Algorithm I:** Shamap
---
**Input**: $N \times N$ Euclidian distance matrix $\mathcal{D}$, angular difference matrix $\Theta$ (calculated by Eq. (1)), reduced dimensionality $p$, # of nearest neighbor $k$ ;
1: $index \leftarrow \text{sort}(\mathcal{D})$ in ascending order along each column and return sorting index
2: $\mathcal{D}[index[k + 2: N]] \leftarrow INF$
3: $\mathcal{D} = \min(\mathcal{D}, \mathcal{D}^t)$ compare matrix element-wise and return smaller elements
4: $\Theta[index[k + 2: N]] \leftarrow INF$
5: $\Theta = \min(\Theta, \Theta^t)$ compare matrix element-wise and return smaller elements
6: **For** $k$ from 1 to $N$ **do**
7:   **For** $i$ from 1 to $N$ **do**
8:     **For** $j$ from 1 to $N$ **do**
9:       **If** $\mathcal{D}[i][j] > \mathcal{D}[i][k] + \mathcal{D}[k][j]$
10:         $\mathcal{D}[i][j] \leftarrow \mathcal{D}[i][k] + \mathcal{D}[k][j]$
11:         $\Theta[i][j] \leftarrow \Theta[i][k] + \Theta[k][j]$
12:       **End if**
13:     **End for**
14:   **End for**
15: **End for**
16: $\mathcal{C} = \cos(\Theta)$
17: $[\lambda, \vec{u}] = \text{eig}(\mathcal{C})$, arrange $\lambda$ and $\vec{u}$ in descending order of $\lambda$
18: Constructing the $p$ dimensional embedding $\mathcal{Y}_p$ using Eq. (3)
**Output**: $\mathcal{Y}_p$

---

In the following, we utilize two examples to illustrate how Shamap decodes data better than Isomap does.

*A. Protein Unfolding*

With advanced Cryo-EM [13], structural biology has made a huge progress over recent years. Determining the structure of a key protein is of great significance for understanding its functions and developing new drugs. In the experiments, it is relatively easy to figure out the configurative amino acid sequence but difficult to estimate the high order structural features [14]. Also, protein unfolding is another challenge. Thus, it is meaningful to describe a protein folding configuration in a low-dimensional space [15].

Here we look at the protein unfolding problem to show the utility of Shamap. The $\alpha$ helices and $\beta$ sheet are very common second order structures of proteins [16]. Without loss of generality, we made a toy protein model consisting of two $\alpha$ helices joint by a $\beta$ sheet, where $\alpha$ helices and $\beta$ sheet are respectively expressed by two tilde spiral equations and a cosine function. Then, we unfolded the protein model with Isomap and Shamap respectively, as shown in Figure 1. Visually speaking, Shamap results are more informative. It is seen that while Isomap mapped the protein into a straight line as expected, Shamap turned the 3D $\alpha$ helices into 2D spirals with the $\beta$ sheet into a straight line in a proper location. In fact, because that Isomap computes the distance along the geodesic line, the yield of Isomap will always be a straight line no matter how curly the protein chain is.

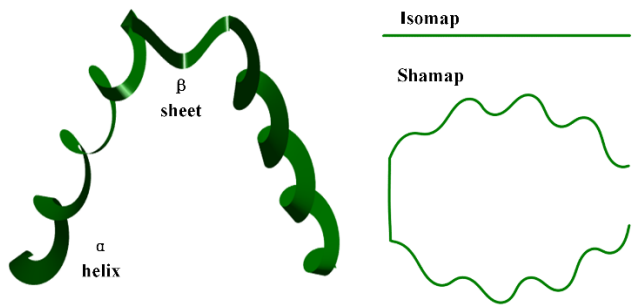

Figure 1. Protein unfolding using Isomap and Shamap respectively.

## B. Digit Orientation

In addition to the above toy examples, we also tested the utility of Shamap with another example: the orientation of "1", where 1000 "1" images were processed to highlight this trend, with the parameters $c = 0, K = 5$ for both Isomap and Shamap. The results are shown in Figure 2. It can be seen that the orientation of "1" projected by Shamap changed gradually along the shape of the fan, with the light and high contrast digits inside and outside respectively. In contrast, the results processed by Isomap just demonstrate the trend of intensity change, totally mixing "1" of different orientation together, which means in Shamap results, similar samples are getting closer and different samples are farther. Therefore, we conclude this example that the features learned by Shamap are more representative.

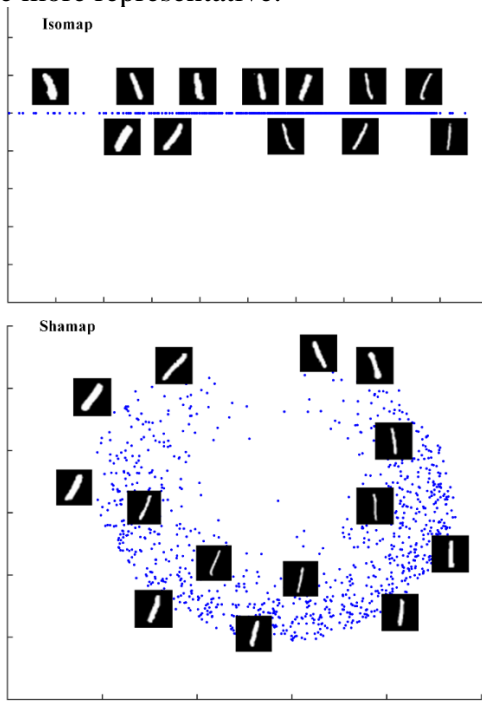

Figure. 2. 1,000 "1" images were analyzed using Shamap.

### III. EXPERIMENTAL RESULTS

To further justify the utility of Shamap, qualitative and quantitative experiments are carried out for clustering and classification with Shamap in comparison to three representative methods: k-PCA, LLE, Isomap. In terms of qualitative assessment, we conducted visualization experiments using a t-distributed stochastic neighbor embedding (t-SNE) projection [17] to provide an understanding of dimension deduction results using different algorithms. t-SNE is one of the most popular data visualization algorithms. For quantitative comparison, clustering and classification are an effective way to test whether data points are in good positions in terms of similarity and difference. An outstanding clustering and classification ability with compressed data would indicate that the latent features are well preserved in a low dimensional space. Here we use K-Means [18] for clustering and SVM [19] for classification of the reduced data cloud.

All experiments were conducted in MATLAB. Four datasets were used in our study, as summarized in TABLE 1. MNIST contains 10,000 28 × 28 gray scale images of digits ranging from 0 to 9. Each digit is in 1,000 images. ORL contains 32 × 32 gray scale images of 40 persons. Each person was in 10 images viewed from different angles. GTFD contains 32 × 32 gray scale face images from 50 persons. Each person was in 15 images with different backgrounds and illumination conditions. CMUPIE contains 32 × 32 gray scale face images of 68 persons. Each person was captured in 42 images under different viewing and illumination conditions.

TABLE I: STATISTICS OF THE THREE DATASETS

| Dataset | Size | Dimensionality | # of classes |
|---------|------|----------------|--------------|
| MNIST   | 10000 | 784           | 10           |
| ORL     | 400  | 1024           | 40           |
| GTFD    | 750  | 1024           | 50           |
| CMUPIE  | 2856 | 1024           | 68           |

In order to measure how accurate our data are clustered in a low dimensional space with regard to the ground truth, we used two rubrics for evaluation. They are accuracy (ACC) and normalized mutual information (NMI) [20]. ACC is measured as the function:

$$ACC = \frac{\sum_{i=1}^{n} \delta(g_i, \text{map}(r_i))}{n} \quad (4)$$

where $g_i$ is the ground truth label, $r_i$ is the cluster label, $n$ is the total number of data, $\delta(x,y)$ is the delta function that equals one if $x = y$ and equals zero otherwise, and map$(x)$ is a function mapping the cluster label to the corresponding true label. The mapping function is computed using the Hungarian algorithm [21]. Hungarian algorithm is a popular combinatorial optimization algorithm which can accurately find the best assignment between the two sides of a bipartite graph. If we formulate the cost edges intro a cost matrix, our purpose will be finding the minimum cost matching. For our clustering results, we set the costs as the negative of number of overlapping labels between the ground truth label vector and the cluster label vector for all kinds of $g$ to $r$ assignments.

NMI is a measurement on mutual dependence between two variables being defined as

$$NMI(C, C^{'}) = \frac{MI(C, C^{'})}{\max(H(C), H(C^{'}))} \quad (5)$$

where $MI(C, C^{'})$ is the mutual information between the ground truth labels and clustered labels, $H(C)$ and $H(C^{'})$ denote the entropies of $C$ and $C^{'}$ respectively. The details of these two matrices can be found in [20].

For evaluation on classification tasks, we use the accuracy rate (ACC), which computes the correspondence between the output of a classifier and the true label:

$$ACC = \frac{\sum_{i=1}^{n} \delta(g_i, r_i)}{n} \tag{6}$$

This is similar to Eq. (5) without the mapping function.

*A. Visualization*

We used the t-SNE projection to visualize the dimensionality reduction effects of Isomap, LLE, k-PCA and Shamap. We conducted the visualization tests on MNIST and CMUPIE datasets that have relatively large cardinality. All of the images were embedded into a 250-dimensional space using these four algorithms [22]. Then, the data dimensionality was further reduced into a 2D space and visualized via t-SNE.

It can be seen in Figures 3 and 4 that k-PCA fails to separate different classes of data in a lower dimensional space. For the MNIST data, the results of LLE, Isomap and Shamap all showed good separability but in different distributions. Shamap presented more clearly on the CMUPIE dataset than the other methods. Nearly all of the 10 classes were separated with little overlapping by Shamap. The LLE results were very good as well, but class 4, represented in light green, was seriously divided in a low dimensional space in comparison to what we had with Shamap, Isomap and k-PCA performed inferiorly, both of which failed to encode the CMUPIE data effectively.

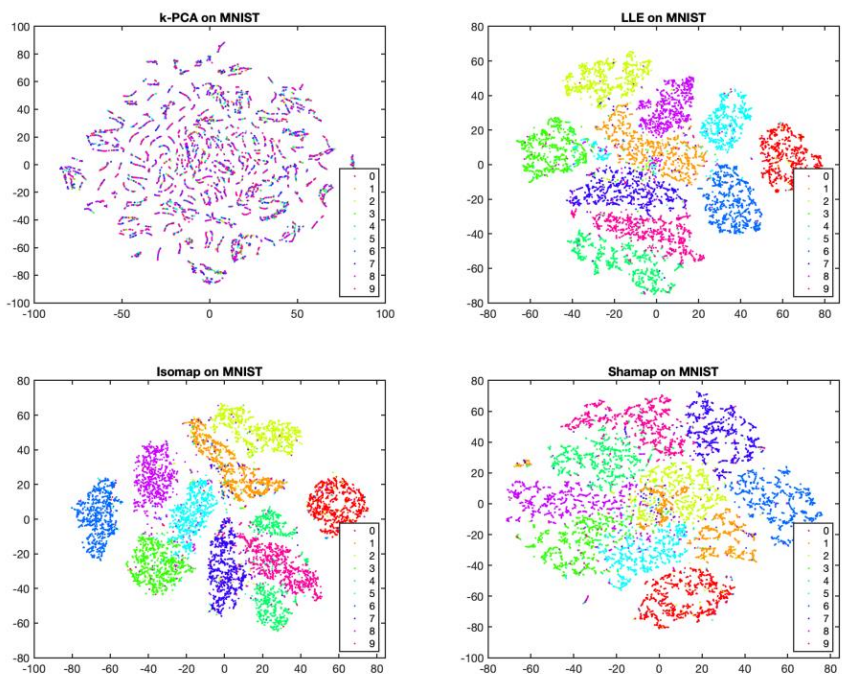

Fig. 3. Visualization of the MNIST data of the first 10 digits. The dimensionality reduction tests were done using **upper left:** k-PCA; **upper right:** LLE; **lower left:** Isomap; **lower right:** Shamp methods.

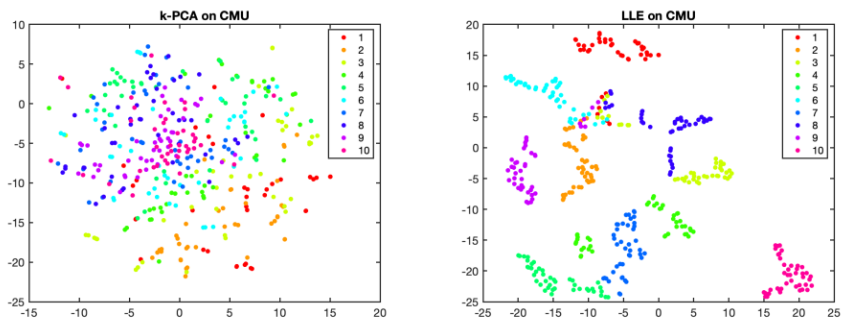

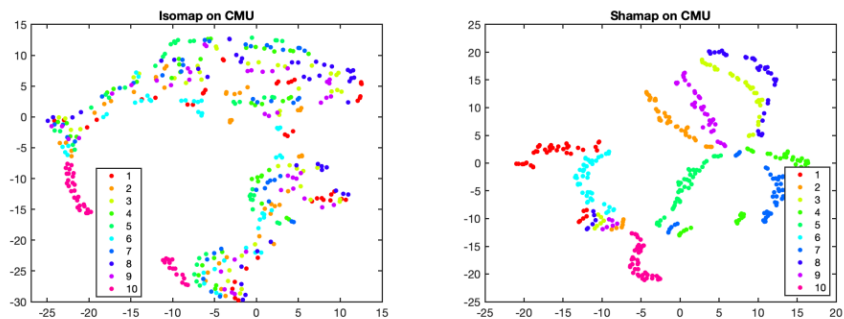

Fig. 4. Visualization of the CMUPIE data of the first 10 categories. The dimensionality reduction tests were done using **upper left:** k-PCA; **upper right:** LLE; **lower left:** Isomap; **lower right:** Shamp methods.

## B. Clustering

In order to better compare the algorithms in the context of clustering, we experimented with data ranging from 2 classes to 10 classes. All of the algorithms (Shamap, LLE and Isomap) used K nearest neighbors, having the same number of neighbors: K=10. The clustering process was repeated 20 times, and the best results (the lowest within-class point-to-centroid; we should measure both inter-class and intra-class distances!) were returned. Figures 5, 6 and 7 demonstrate the performance of the algorithms on different numbers of classes. For the ORL dataset, the clustering ability of Shamap is slightly better than that of Isomap, but both Isomap and Shamap are significantly better than LLE and k-PCA in all the cases. For the GTFD dataset, Isomap and Shamap performed similarly, being superior to LLE and k-PCA. For the CMUPIE dataset, LLE and Shamap took the leading position over Isomap and k-PCA, being implied by the above-mentioned visualization results. Overall, the performance of LLE and Shamap are comparable in this test. The NMI and ACC scores of the four algorithms on three benchmarks are summarized in TABLE II. The highest scores with respect to different neighbors are bolded. It can be concluded that Shamap delivers competitive clustering performance compared to other algorithms.

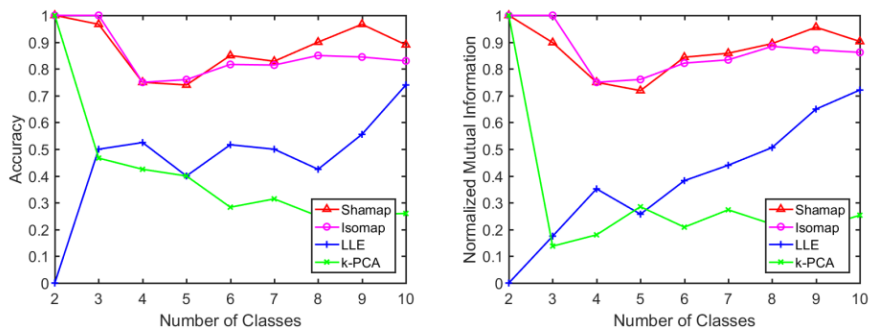

Fig. 5. Clustering performance of 4 algorithms on the ORL dataset: **left:** The accuracy vs. number of classes; and **right:** NMI vs. number of classes.

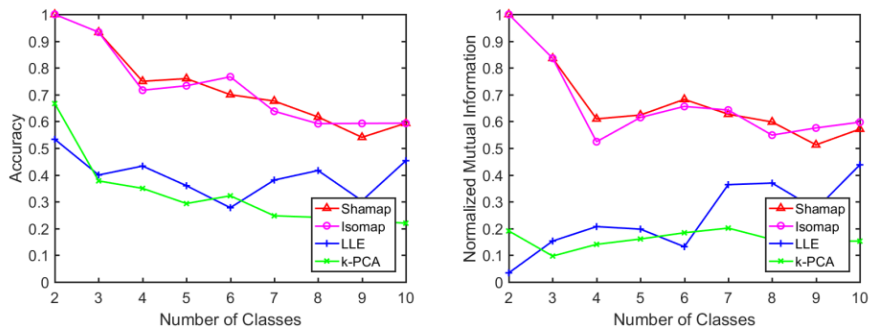

Fig. 6. Clustering performance of 4 algorithms on the GTFD dataset: **left:** The accuracy vs. number of classes; and **right:** NMI vs. number of classes.

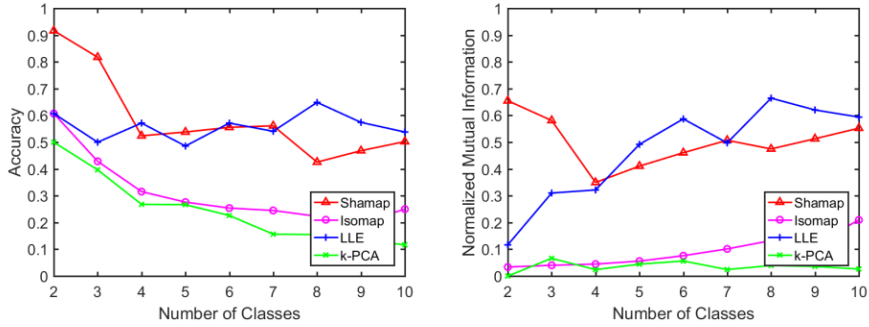

Fig. 7. Clustering performance of 4 algorithms on the CMUPIE dataset: **left:** The accuracy vs. number of classes; and **right:** NMI vs. number of classes.

TABLE II: CLUSTERING PERFORMANCE OF SHAMAP, ISOMAP, LLE, K-PCA ON THE ORL, GTFD AND CMUPIE DATASETS RESPECTIVELY.

| Dataset | Metrix | Algorithm | $K=8$ | $K=9$ | $K=10$ | $K=11$ | $K=12$ | $K=13$ | $K=14$ | $K=15$ |
|---|---|---|---|---|---|---|---|---|---|---|
| ORL | ACC | Shamap | **0.96** | **0.8** | 0.74 | **0.76** | **0.8** | **0.78** | **0.8** | 0.72 |
| | | LLE | 0.52 | 0.52 | 0.4 | 0.48 | 0.34 | 0.42 | 0.48 | 0.56 |
| | | Isomap | / | / | **0.76** | 0.76 | 0.8 | 0.78 | 0.78 | **0.8** |
| | | k-PCA | 0.44 | 0.34 | 0.4 | 0.34 | 0.34 | 0.34 | 0.38 | 0.38 |
| | NMI | Shamap | **0.917** | **0.749** | 0.718 | 0.72 | **0.787** | **0.745** | **0.787** | 0.752 |
| | | LLE | 0.3821 | 0.321 | 0.257 | 0.323 | 0.202 | 0.271 | 0.402 | 0.359 |
| | | Isomap | / | / | **0.761** | **0.761** | 0.731 | 0.745 | 0.786 | **0.787** |
| | | k-PCA | 0.327 | 0.193 | 0.285 | 0.19 | 0.19 | 0.193 | 0.183 | 0.234 |
| GTFD | ACC | Shamap | **0.76** | **0.733** | **0.76** | **0.706** | **0.666** | **0.666** | 0.64 | **0.653** |
| | | LLE | 0.4 | 0.266 | 0.36 | 0.373 | 0.453 | 0.466 | 0.493 | 0.44 |
| | | Isomap | 0.68 | 0.733 | 0.733 | 0.64 | 0.666 | 0.666 | **0.68** | 0.64 |
| | | k-PCA | 0.373 | 0.386 | 0.293 | 0.293 | 0.373 | 0.293 | 0.333 | 0.293 |
| | NMI | Shamap | **0.623** | **0.616** | **0.623** | **0.639** | **0.589** | **0.61** | 0.589 | **0.578** |
| | | LLE | 0.242 | 0.112 | 0.197 | 0.2 | 0.262 | 0.36 | 0.303 | 0.228 |
| | | Isomap | 0.523 | 0.593 | 0.614 | 0.597 | 0.61 | 0.592 | **0.591** | 0.563 |
| | | k-PCA | 0.182 | 0.195 | 0.161 | 0.174 | 0.212 | 0.159 | 0.171 | 0.164 |
| CMUPIE | ACC | Shamap | **0.628** | **0.614** | 0.538 | 0.49 | 0.485 | 0.533 | 0.495 | 0.49 |
| | | LLE | 0.623 | 0.48 | **0.542** | **0.538** | **0.623** | **0.628** | **0.804** | **0.585** |
| | | Isomap | 0.29 | 0.271 | 0.276 | 0.271 | 0.276 | 0.266 | 0.266 | 0.276 |
| | | k-PCA | 0.223 | 0.219 | 0.266 | 0.223 | 0.252 | 0.219 | 0.276 | 0.223 |
| | NMI | Shamap | **0.493** | **0.506** | 0.41 | 0.372 | 0.368 | 0.393 | 0.365 | 0.36 |
| | | LLE | 0.493 | 0.44 | **0.491** | **0.52** | **0.658** | **0.637** | **0.74** | **0.539** |
| | | Isomap | 0.082 | 0.052 | 0.055 | 0.052 | 0.052 | 0.047 | 0.041 | 0.053 |
| | | k-PCA | 0.027 | 0.019 | 0.044 | 0.026 | 0.029 | 0.023 | 0.048 | 0.029 |

("/" is the result of that Isomap cannot find short paths for some points when $K=8,9$ for ORL dataset)

## C. Classification

Similarly, we conducted the classification tests with data ranging from 5 classes to 10 classes. All of the algorithms involved K-Nearest Neighbors (Shamap, LLE and Isomap). The classification algorithm SVM was used. Figure 8 shows the classification results of the four algorithms on the ORL and GTFD datasets. It is observed that Shamap offered a competitive performance on the ORL and GTFD datasets. LLE, Shamap and Isomap outperformed k-PCA in a large margin on the ORL benchmark, while on the GTFD dataset, LLE won the first place from six to ten classes immediately followed by Shamap.

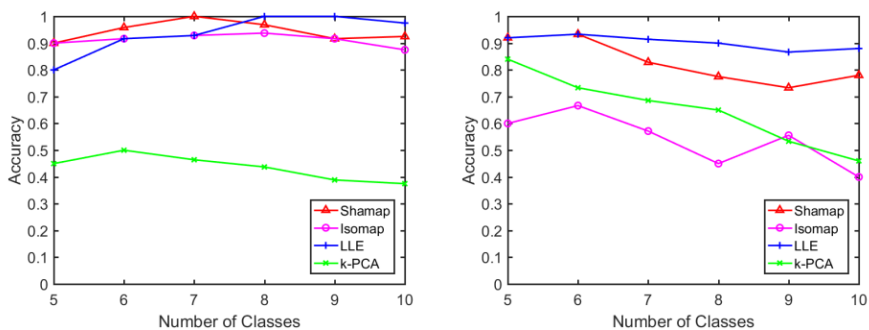

Fig. 8. Classification accuracy vs. number of classes of 4 algorithms on three datasets: **left:** The accuracy vs. number

TABLE III: CLASSIFICATION PERFORMANCE OF SHAMAP, ISOMAP, LLE, K-PCA ON ORL, GTFD AND CMUPIE DATASET

| Dataset | Algorithm | $K = 8$ | $K = 9$ | $K = 10$ | $K = 11$ | $K = 12$ | $K = 13$ | $K = 14$ | $K = 15$ |
|---|---|---|---|---|---|---|---|---|---|
| ORL | Shamap | 0.825 | 0.775 | 0.85 | 0.825 | 0.925 | 0.925 | **1** | 0.95 |
|  | Isomap | / | / | 0.85 | 0.775 | 0.875 | 0.9 | 0.975 | 0.875 |
|  | LLE | **0.975** | **0.975** | **0.975** | **0.975** | **0.975** | **0.975** | 0.975 | **0.975** |
|  | KPCA | 0.375 | 0.375 | 0.375 | 0.375 | 0.375 | 0.375 | 0.375 | 0.375 |
| GTFD | Shamap | 0.76 | 0.78 | 0.8 | 0.78 | 0.78 | 0.78 | 0.76 | 0.74 |
|  | Isomap | 0.4 | 0.42 | 0.38 | 0.44 | 0.4 | 0.42 | 0.4 | 0.38 |
|  | LLE | **0.84** | **0.86** | **0.86** | **0.88** | **0.88** | **0.88** | **0.86** | **0.88** |
|  | KPCA | 0.46 | 0.46 | 0.46 | 0.46 | 0.46 | 0.46 | 0.46 | 0.46 |

("/" is the result of that Isomap cannot find short paths for some points when $K = 8,9$ for ORL dataset)

of classes on ORL dataset; and **right:** accuracy vs. number of classes on GTFD dataset.

### D. Robustness to K

Robustness on different parameters is important for an algorithm. The parameter $K$ (the number of nearest neighbors) is the key parameter of Shamap, Isomap and LLE. Hence, we evaluated the influence of $K$ on the clustering performance of the above three algorithms. Figure 9 shows the results with these algorithms versus $K$ ranging from 8 to 15. It is found that Shamap was robust with respect to different $K$ values and worked well on all datasets. Interestingly, Isomap is widely known to be sensitive to $K$ [23]. For example, Isomap could not find effective geodesic distance between some data points of ORL when $K$=8 and $K$=9 so that it failed to reduce the data dimensionality well, as shown in Figure 6. Although LLE outperformed Shamap on the CMUPIE dataset, it performed poorly on the other two datasets. As KNN is not involved in k-PCA, we did not include it in this comparison.

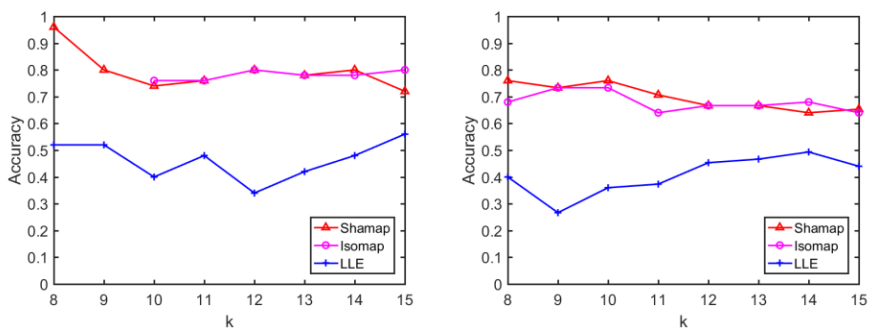

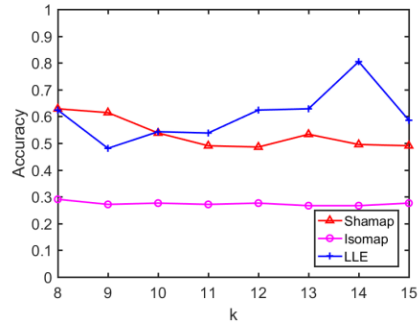

Fig. 9. Clustering performance of 3 algorithms versus different K values on the three datasets: **upper left:** The accuracy vs. K on ORL dataset; **upper right:** Accuracy vs. k on the GTFD dataset; **lower:** Accuracy vs. k on the CMUPIE dataset.

## IV. Discussions and Conclusion

In a neighborhood of each point, Isomap computes the geodesic distance between the neighboring points on a manifold. In contrast, Shamap measures the accumulated angular change along the geodesic line with respect to a reference point; i.e., angles on the surface of the manifold are calculated. Therefore, Shamap is good at keeping shape information. As illustrated in Figure 2, while Isomap tends to unfold proteins into a straight line, Shamap carries shape differences from a high--dimensional space to a low-dimensional one. An interesting topic is to perform a topology-preserving dimensionality reduction. We believe that Shamap is a step forward from Isomap towards this goal.

In conclusion, we have proposed a new nonlinear algorithm, Shamap, for dimensionality reduction. The main merit of Shamap is its shape preserving property. Our pilot studies show shape-information-rich results after dimensionality reduction using Shamap, favorably compared with the counterparts obtained using Isomap. Further efforts are in progress to apply Shamap in real-world applications and improve it for topological invariability.